# Insuring Smiles: Predicting routine dental coverage using Spark ML

Aishwarya Gupta, Rahul S. Bhogale, Priyanka Thota, Prathushkumar Dathuri, Jongwook Woo
Department of Information Systems, California State University Los Angeles
{agupta25, rbhogal, pthota2, pdathur, jwoo5}@calstatela.edu

**Abstract:** Finding suitable health insurance coverage can be challenging for individuals and small enterprises in the USA. The Health Insurance Exchange Public Use Files (Exchange PUFs) dataset provided by CMS offers valuable information on health and dental policies [1]. In this paper, we leverage machine learning algorithms to predict if a health insurance plan covers routine dental services for adults. By analyzing plan type, region, deductibles, out-of-pocket maximums, and copayments, we employ Logistic Regression, Decision Tree, Random Forest, Gradient Boost, Factorization Model and Support Vector Machine algorithms. Our goal is to provide a clinical strategy for individuals and families to select the most suitable insurance plan based on income and expenses.

## 1. Introduction

Our work focuses on using machine learning algorithms to predict dental coverage in health insurance plans. We chose this topic due to the challenge individuals and small enterprises face in finding suitable health insurance coverage. By leveraging the CMS dataset [1], we aim to provide insights on routine dental service coverage based on plan details and other relevant factors.

The importance of our work lies in supporting timely benefits analysis, rate evaluation, copayment assessment, premium analysis, and geographic coverage analysis. By applying machine learning algorithms such as Logistic Regression, Decision Tree, Random Forest, Support Vector Machine, Gradient Boost and Factorization Machine we contribute to the field by offering practical recommendations for selecting appropriate health insurance plans based on income and expenses.

Our background involves utilizing big data technologies and advanced machine learning techniques to improve classification accuracy. By comparing the performance of different algorithms, we provide valuable insights for individuals and insurance stakeholders in making informed decisions regarding health coverage options.

## 2. Related Work

In one of the similar works found in the Hamilton Project, smart policies on health insurance were developed using the Healthcare.gov dataset [1]. The study aimed to create a platform for suggesting smart policies based on cost reductions and insurance coverage offered by employers. In contrast, our analysis focuses on predicting dental coverage in health insurance plans and determining if routine dental services are included. Additionally, our approach involves utilizing big data file management, HDFS, along with PySpark ML using Databricks/Zeppelin notebooks.

Another related work was presented by the NAIC (National Association of Insurance Commissioner) [3]. Their study primarily focuses on the business side of Health insurance marketplace data and analyzes the growth in issuing plans over a 10-year period. They observe a decrease in net earnings and profit margin. In contrast, our work focuses on providing insights and predictions related to routine health and dental checkup services, with a focus on consumer needs.

The work presented by the AMA (American Medical Association) emphasizes health insurance companies with wider market coverage and explores metrics such as enrollment% and small groups% [4]. In contrast, our analysis differs in terms of deliverables as we concentrate on individual/family plan needs. We compare various machine learning models to gain insights into different insurance plans, specifically regarding dental coverage and routine services.

The key difference in our work lies in the utilization of big data technologies such as Hadoop, PySpark ML, and cloud computing. Our analysis leverages the power of big data processing and analytics to predict dental coverage in health insurance plans, offering valuable insights for individuals and small enterprises in selecting the most suitable insurance options.

## 3. Background Work

Our paper analyzes health insurance coverage and utilization of health services in 38 Sub-Saharan African countries using recent Demographic Health Survey (DHS) data. The study reveals low enrolment rates and utilization of health services in the region. Machine learning (ML) techniques significantly improve the accuracy of predicting health insurance coverage compared to traditional approaches. The findings underscore the need to identify and include the excluded population and highlight the importance of country specific factors in predicting enrolment. The research demonstrates the potential of ML models to target policies and improve overall targeting of the excluded population. Efforts to achieve Universal Health Coverage (UHC) should focus on improving coverage and utilization rates. [5]

## 4. Specification

The dataset comprises of insurance plans related to planning rates, types, coverage, rules, family members, etc. The dataset is of size 2.1 GB having a duration from 2017 to 2021 [1]. Table 1 shows numerous files, file types, and their size of it.

*Table 1 Data Specification*

| Data Set | Size (Total 2.1 GB) |
|---|---|
| benefits_cost_sharing.csv | 2,050,850 KB |
| benefits1.csv (Sampled dataset) | 12,901 KB |

Table 2 shows the hardware specifications for the Hadoop cluster and CPU configuration.

*Table 2 H/W Specification*

| | |
|---|---|
| **Hadoop Specs** | |
| Master Nodes | 2 |
| Worker Nodes | 3 |
| Version | 3.1.2 |
| Storage | 390.7 GB |
| **CPU Configuration** | |
| CPU core | 6 |
| Frequency | 1995.312 MHz |

## 5. Implementation Flowchart

The outlined flowchart in Figure 1 illustrates the process of analyzing a dataset obtained from CMS gov website, beginning with acquiring the dataset and uploading it to a Linux system, storing it in HDFS. The next steps involve loading the dataset, performing initial data exploration, handling missing values, and removing irrelevant columns. If necessary, data transformation and normalization are applied, followed by data quality checks.

A modeling pipeline is defined, encompassing tasks such as data splitting, feature engineering, data scaling, and encoding categorical variables. Models are built by selecting algorithms, training them, and saving the trained models. Evaluation metrics are used to compare algorithms for their accuracy and performance. Prediction is applied to the testing set, evaluating performance, and comparing their effectiveness. The flowchart serves as a guide for a structured and comprehensive approach to the data analysis process in Figure 1.

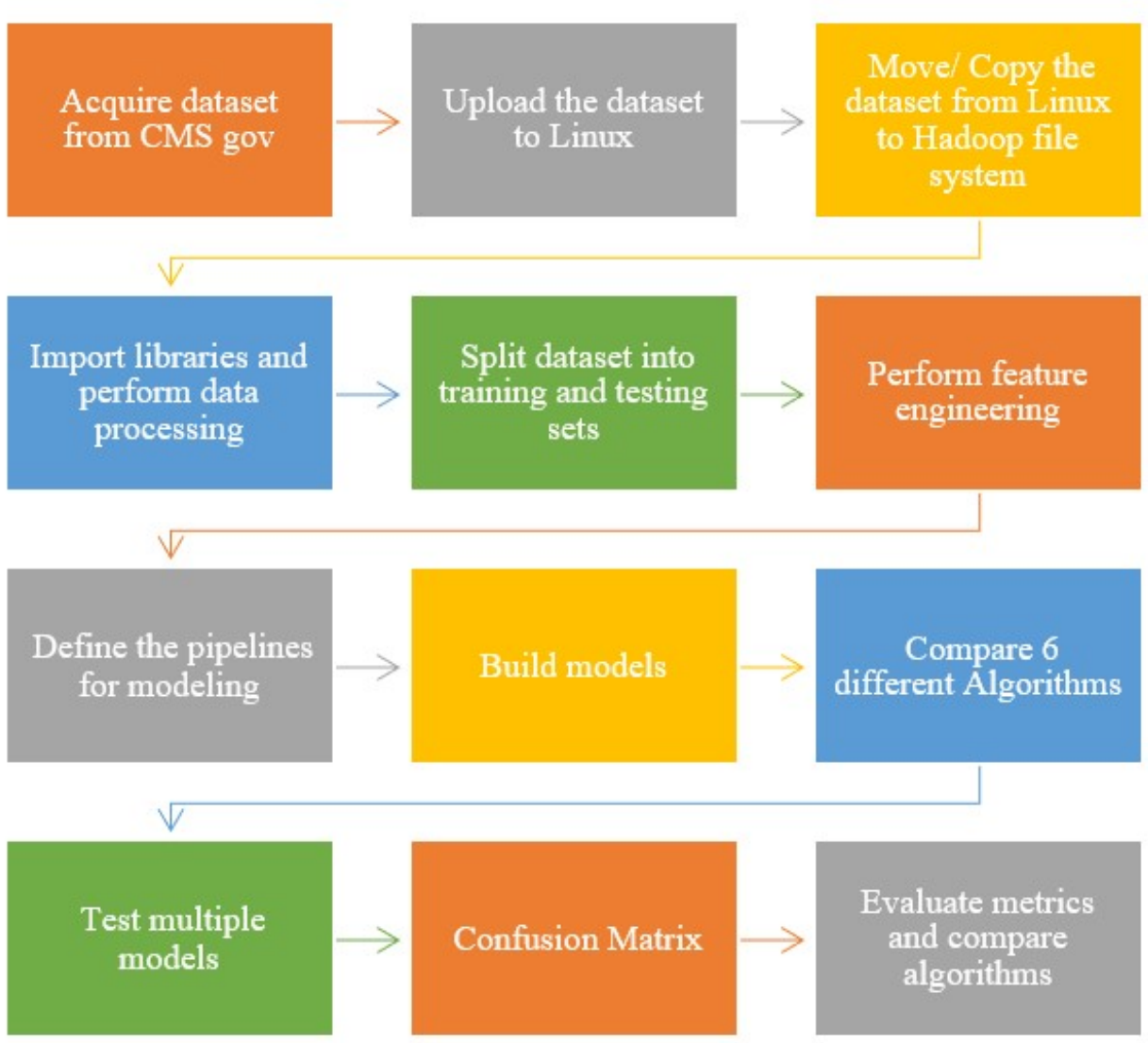

*Figure 1 Implementation Flowchart*

## 6. Data Preprocessing

As part of the data preparation process before building the pipeline, we needed to convert categorical data into numerical data since the dataset contained nonnumeric values. Since the dataset was balanced, with 60% of health insurance plans covering dental services and the remaining 40% not covering dental services, we opted to use classification algorithms.

For testing purposes, we randomly sampled only 50,000 rows in by using Python code in a Jupyter notebook in Figure 2, and later loaded into Databricks platform in Figure 3. We selected independent columns for our prediction, including [BusinessYear, StateCode, IssuerId, SourceName, IsEHB, QuantLimitOnSvc, Exclusions, EHBVarReason, IsCovered], to ensure unbiased predictions.

Next, we identified and removed null values only from the prediction column in Figure. The remaining null values in other columns were replaced with the maximum value from each respective column in Figure 5. Later convert the 'IsCovered' column into 0 and 1 for classification modelling as this is our prediction column in Figure 6.
After preprocessing the dataset, we split it into a 70% training set and a 30% testing set for modeling purposes.

import pandas as pd
import random
# Load the original dataset file into a DataFrame
df = pd.read_csv('C:\\Users\\agupta25\\Desktop\\5560 project\\benefits_cost_sharing.csv')
# Randomly select 10,000 rows from the DataFrame
num_rows = min(50000, len(df))  # Select either 50,000 rows or the total number of rows if it's less than 50,000
random.seed(42)  # Set a seed for reproducibility
indices = random.sample(range(len(df)), num_rows)
df_sample = df.loc[indices]
# Save the selected rows as a new CSV file
df_sample.to_csv('benefits1.csv', index=False)
df_sample

*Figure 2 – Sampled dataset using Python*

###Load the sample dataset from DBFS
# Oracle BDCE
#csv = spark.read.csv('/user/agupta25/project/benefits1.csv', inferSchema=True, header=True)
# File location and type
file_location = "/FileStore/tables/benefits1.csv"
file_type = "csv"
# CSV options
infer_schema = "true"
first_row_is_header = "true"
delimiter = ","
df = spark.read.format(file_type) \
.option("inferSchema", infer_schema) \
.option("header", first_row_is_header) \
.option("sep", delimiter) \
.load(file_location)
display(df)

*Figure 3 – Loaded sampled csv in Databricks*

###Replace null or whitespace values with None. Later drop the values.
from pyspark.sql.functions import when, col
# Replace empty strings or whitespace with null values
df = df.withColumn('label', when(col('label').isin('', ' '), None).otherwise(col('label')))
# Drop null values from label column
df = df.dropna(subset=['label'])
df.show()

*Figure 4 – Drop the null value of prediction column*

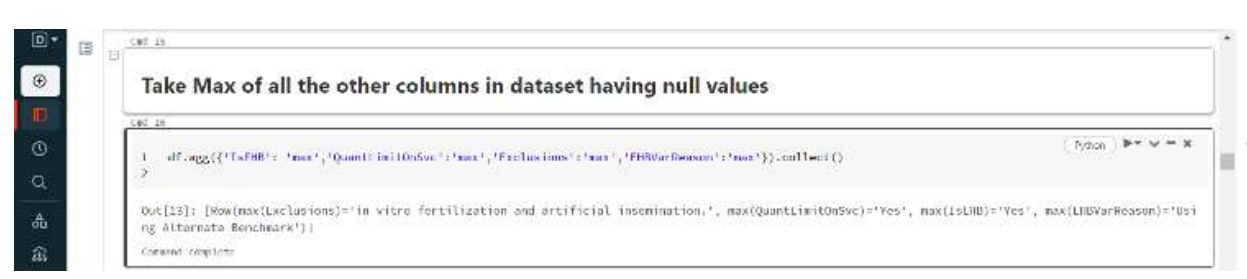

*Figure 5 Replace null values of other columns with their maximum*

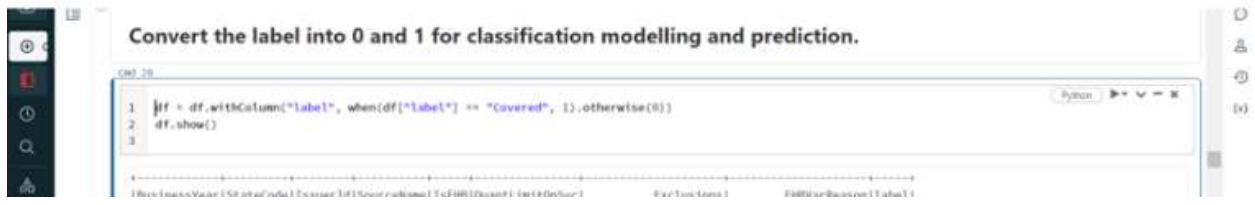

*Figure 6 Labelled into 0's and 1's*

## 7. Defining Pipeline

- A pipeline consists of a series of transformer and estimator stages that typically prepare a DataFrame for
- modeling and then train a predictive model. In this case, you will create a pipeline with seven stages:
- A StringIndexer estimator that converts string values to indexes for categorical features.
- A VectorAssembler that combines categorical features into a single vector.
- A VectorIndexer that creates indexes for a vector of categorical features.
- A VectorAssembler that creates a vector of continuous numeric features.
- A MinMaxScaler that normalizes continuous numeric features.
- A VectorAssembler that creates a vector of categorical and continuous features.

A predictive model often requires multiple stages of feature preparation. For example, it is common when using some algorithms to distinguish between continuous features (which have a calculable numeric value) and categorical features (which are numeric representations of discrete categories). It is also common to normalize continuous numeric features to use a common scale (for example, by scaling all numbers to a proportional decimal value between 0 and 1) in Figure 7.

```
strIdx_SC = StringIndexer(inputCol = "StateCode", outputCol = "SC",handleInvalid='keep')
strIdx_SN = StringIndexer(inputCol = "SourceName", outputCol = "SN",handleInvalid='keep')
strIdx_EHB = StringIndexer(inputCol = "IsEHB", outputCol = "EHB",handleInvalid='keep')
strIdx_QL = StringIndexer(inputCol = "QuantLimitOnSvc", outputCol = "QL",handleInvalid='keep')
strIdx_EX = StringIndexer(inputCol = "Exclusions", outputCol = "EX",handleInvalid='keep')
strIdx_EHBVR = StringIndexer(inputCol = "EHBVarReason", outputCol = "EHBVR",handleInvalid='keep')

# the following columns are categorical number such as ID so that it should be Category features
catVect = VectorAssembler(inputCols = ["SC", "BusinessYear", "IssuerId", "SN", "EHB","QL","EHBVR"], outputCol="catFeatures")
catIdx = VectorIndexer(inputCol = catVect.getOutputCol(), outputCol = "idxCatFeatures", handleInvalid="skip")

# cat feature vector is normalized

minMax = MinMaxScaler(inputCol = catIdx.getOutputCol(), outputCol="normFeatures")

featVect = VectorAssembler(inputCols=["normFeatures"], outputCol="features")
```

*Figure 7 – Defining Pipeline*

## 8. Training a Model

In our analysis, we utilize six ML algorithms to predict dental coverage in health insurance plans. These algorithms are Logistic Regression (LR), Decision Tree (DT), Random Forest (RF), Factorization Machine (FM), Gradient Boost (GBT), and Support Vector Machine (SVM) in Figure 8.
To optimize their performance, we tune the hyperparameters using a parameter grid in Figure 9.

```
classification_models=["Logistic Regression (LR)","Decision Tree (DT)","Random Forest (RT)","Factorization Machine (FM)","Gradiest Boost (GBT)","Sup

#creating diff clasf algos for testing accuracy,computing time, precision, recall, ROC, PR
cls_mod=[]

cls_mod.insert(0,LogisticRegression(labelCol="label",featuresCol="features",maxIter=10,regParam=0.3,threshold=0.35))
cls_mod.insert(1,DecisionTreeClassifier(labelCol="label", featuresCol="features",seed=42))
cls_mod.insert(2,RandomForestClassifier(labelCol='label', featuresCol='features',seed=42))
cls_mod.insert(3,FMClassifier(labelCol='label', featuresCol='features', seed=42))
cls_mod.insert(4,GBTClassifier(labelCol='label', featuresCol='features', seed=42))
cls_mod.insert(5,LinearSVC(labelCol='label', featuresCol='features'))

# define list of models made from Train Validation Split or Cross Validation
model = []
pipeline = []

# Pipeline process the series of transformation above, which is another transformation
for i in range(0,6):
    pipeline.insert(i,Pipeline(stages=[strIdx_SC,strIdx_SN,strIdx_EHB,strIdx_QL,strIdx_EHBVR, catVect, catIdx,minMax, featVect, cls_mod[i]]))
```

*Figure 8 – Defining Models*

For evaluation, we employ a Cross-Validator model that compares the predicted values with the true labels (IsCovered column) using a binary classification evaluator in Figure 10. Metrics like accuracy, precision, and recall are used to assess these algorithms' performance.

```
paramGrid.insert(5,ParamGridBuilder() \
            .addGrid(cls_mod[5].regParam, [0.01, 0.5]) \
            .addGrid(cls_mod[5].maxIter, [1, 5]) \
            .addGrid(cls_mod[5].tol, [1e-4, 1e-3]) \
            .addGrid(cls_mod[5].fitIntercept, [True, False]) \
            .addGrid(cls_mod[5].standardization, [True, False]) \
            .build())
```

*Figure 9 – Parameter building*

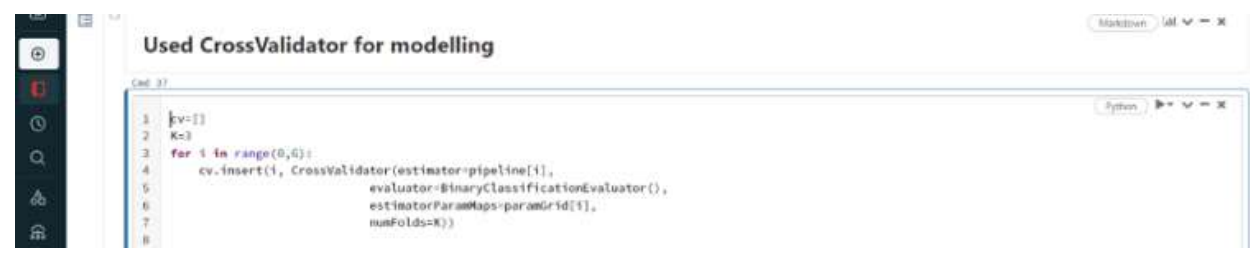

*Figure 10 – Training Model*

### 8.1 Testing

The trained model, which consists of all the stages in the pipeline, is a transformer that can be applied to a specific DataFrame to generate predictions. This model is used to verify the predictions on the test dataset, which represents 30% of the actual dataset. The resulting DataFrame contains the predicted values in the prediction column and the actual known values in the trueLabel column in Figures 11.1 & 11.2

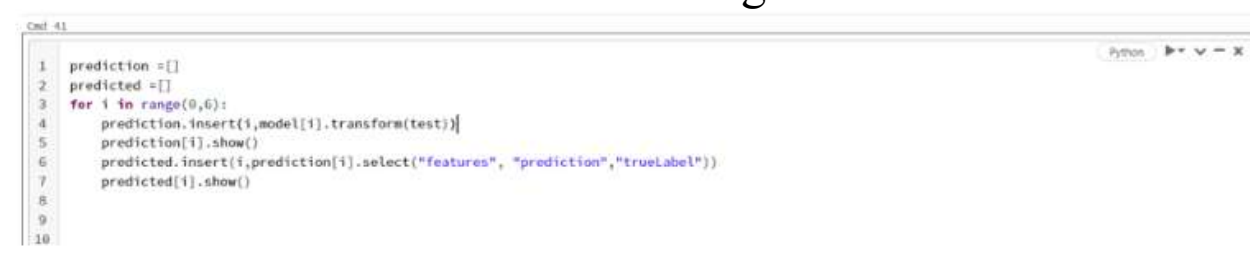

*Figure 11.1 – Testing Model*

```
+--------------------+----------+---------+
|            features|prediction|trueLabel|
+--------------------+----------+---------+
|[1.0,0.0,0.709384...|       1.0|        1|
|(7,[0,2,4],[1.0,0...|       1.0|        1|
|[0.89473684210526...|       1.0|        1|
|[0.89473684210526...|       1.0|        1|
|[0.89473684210526...|       1.0|        1|
|[0.89473684210526...|       1.0|        1|
|[0.89473684210526...|       1.0|        0|
|[0.65789473684210...|       1.0|        1|
|[0.65789473684210...|       1.0|        1|
|[0.65789473684210...|       1.0|        1|
|[0.65789473684210...|       1.0|        1|
|[0.65789473684210...|       1.0|        1|
|[0.65789473684210...|       1.0|        1|
|[0.65789473684210...|       1.0|        0|
|[0.65789473684210...|       1.0|        0|
|[0.21052631578947...|       1.0|        1|
|(7,[0,2,4],[0.210...|       1.0|        1|
|(7,[0,2,4],[0.210...|       1.0|        1|
```

*Figure 11.2 – Prediction and true label on test dataset*

### 8.2. Evaluation & Prediction

The evaluation metrics used for performance assessment of the ML algorithms are as follows:

**Computing time:** Time required for computations and running ML algorithms on the dataset in Figure 12.
**Confusion matrix:** Matrix indicating true positives, true negatives, false positives, and false negatives.
**Precision:** Proportion of correctly predicted positive instances out of the total instances predicted as positive.
**Recall:** Proportion of correctly predicted positive instances out of all actual positive instances in Figure 13.
**ROC (Receiver Operating Characteristic):** Graphical representation of the model's performance by plotting true positive rate against false positive rate.
**PR (Precision-Recall):** Graphical representation of precision against recall at different classification thresholds.
**Accuracy:** Ratio of correctly classified instances to the total number of instances.

**F1 Score:** Harmonic mean of precision and recall, providing a balanced measure of the model's performance.

These metrics help assess the accuracy, sensitivity, specificity, and overall correctness of the models' predictions.

```
Cmd 39

import time

start_time = []
end_time = []
computation_time = []

for i in range(0, 6):
    start_time.insert(i, time.time())
    model.insert(i, cv[i].fit(train))
    # model1 = cv1.fit(train)
    # model2 = cv2.fit(train)
    # model3 = cv3.fit(train)
    # model4 = cv4.fit(train)
    # model5 = cv5.fit(train)
    end_time.insert(i, time.time())
    computation_time.insert(i, (end_time[i] - start_time[i]) / 60.0)
    print("Computation time:",i," ",computation_time[i], "minutes")

/databricks/spark/python/pyspark/ml/util.py:886: UserWarning: Cannot find mlflow module. To enable MLflow logging, install mlflow from PyPI.
  warnings.warn(_MLflowInstrumentation._NO_MLFLOW_WARNING)
Computation time: 0   5.152820577589671 minutes
Computation time: 1   29.827600558598835 minutes
Computation time: 2   36.10357682307561 minutes
Computation time: 3   13.262092169125875 minutes
Computation time: 4   6.539898025989532 minutes
Computation time: 5   15.529444551467895 minutes
Command complete
```

*Figure 12 – Computing time*

```
Cmd 45

for i in range(0,6):
    tp = float(predicted[i].filter("prediction== 1.0 AND truelabel == 1").count())
    fp = float(predicted[i].filter("prediction== 1.0 AND truelabel == 0").count())
    tn = float(predicted[i].filter("prediction== 0.0 AND truelabel == 0").count())
    fn = float(predicted[i].filter("prediction==0.0 AND truelabel == 1").count())
    precision.insert(i,tp / (tp + fp))
    recall.insert(i,tp / (tp + fn))
    metrics.insert(i, spark.createDataFrame([
    ("TP", tp),
    ("FP", fp),
    ("TN", tn),
    ("FN", fn),
    ("Precision", tp / (tp + fp)),
    ("Recall", tp / (tp + fn))],["metric", "value"]))
    metrics[i].show()
```

```
+---------+------------------+
|   metric|             value|
+---------+------------------+
|       TP|           11699.0|
|       FP|            2713.0|
|       TN|               0.0|
|       FN|               3.0|
|Precision|0.8117540938107133|
|   Recall|0.9997436335669116|
+---------+------------------+
```

*Figure 13 – Confusion Matrix*

## 9. Conclusion

From the experimental results, we conclude that GBT is the best model with highest precision of 85% with least computation time 6.5 mins for the **sampled dataset** in Table 3. The best model is Gradient Boost with computation time of 40 minutes and accuracy with 85% with **actual dataset** in Table 4.

*Table 3 Performance Comparison with the Sample Data*

| **Model** | **Comp. Time (mins)** | **Precision (%)** | **Recall (%)** | **ROC** | **AUC PR** |
|---|---|---|---|---|---|
| LR | 5 | 81 | 0.99 | 0.62 | 0.88 |
| DT | 30 | 87 | 0.99 | 0.65 | 0.85 |
| RF | 36 | 86 | 0.99 | 0.85 | 0.96 |
| FM | 13 | 81 | 0.99 | 0.63 | 0.89 |
| GBT | 6.5 | 85 | 0.99 | 0.83 | 0.95 |
| SVM | 15.5 | 81 | 1.00 | 0.60 | 0.87 |

*Table 4 Performance Comparison with the Entire Data*

| **Model** | **Comp Time (mins)** | **Precision** | **Recall** | **AUC ROC** | **AUC PR** |
|---|---|---|---|---|---|
| LR | 20.5 | 0.81 | 0.99 | 0.62 | 0.88 |
| GBT | 40 | 0.85 | 0.99 | 0.83 | 0.95 |
| FM | 5 | 0.81 | 0.99 | 0.63 | 0.88 |
| SVM | 4 | 0.81 | 1.00 | 0.60 | 0.87 |

Feature Importance in Table 5 is the ranking of *df* columns and observing which columns are impacting the most for predicting whether the dental services are covered or not in health insurance plans based on the value. Thus, we evaluated only for **GBT Classifier**. ***Exclusion*** column has the highest impact with 0.55 compared to ***isEHB*** which is redundant column.

**Table 5. Feature Importance:**

| Ranking | df columns | Importance value |
|---|---|---|
| 1 | Exclusions | 0.555733 |
| 2 | BusinessYear | 0.162801 |
| 3 | IssuerId | 0.131934 |
| 4 | QuantLimitOnSvc | 0.12204 |
| 5 | SourceName | 0.015729 |
| 6 | StateCode | 0.011764 |
| 7 | IsEHB | 0 |